\documentclass[letterpaper, 10 pt, conference]{ieeeconf}  

\usepackage{xcolor}

\usepackage{fancyhdr}
\fancypagestyle{firstpage}{
  \fancyhf{}
  
  \chead{
    \raisebox{1.5cm}[0pt][0pt]{
      \color{gray} 
      \begin{tabular}{c}
        \small This work has been accepted for publication at the \\
        \small IEEE/RSJ International Conference on Intelligent Robots and Systems (IROS), 2026, \copyright IEEE
      \end{tabular}
    }
  }
}

\usepackage{amssymb}
\usepackage{amsmath}
\usepackage{graphicx}
\usepackage{algorithm}
\usepackage{algorithmic}
\usepackage{indentfirst}
\usepackage{dsfont}
\usepackage{subfigure}
\usepackage{booktabs}
\usepackage{multirow}
\usepackage{bm}
\usepackage{tabularx}

\usepackage{amsthm}
\usepackage{afterpage}

\usepackage{graphicx}
\usepackage{tikz}
\usetikzlibrary{arrows.meta,positioning,fit,calc,backgrounds}

\IEEEoverridecommandlockouts

\title{\LARGE \bf
Scaling Learning-based AEB with Massive Unlabeled Data
}

\author{Xiangyu Wang, Yang Zhan, Mengxiang Hao, Chuanchuan Zhong, Yansong Jia, \\ Junjie Zhang, 
Yu Han, Xin Jiang, Zhen Cao, Ying Wang, Yulun Song, and Zhitao Xu
\thanks{All authors are with Li Auto. ({\tt\small wangxiangyu6@lixiang.com})}%
}

\begin{document}

\maketitle
\thispagestyle{firstpage} 
\pagestyle{empty}

\begin{abstract}
This paper studies how to scale learning-based automatic emergency braking (AEB) with massive unlabeled
fleet data under production constraints. Our approach is based on meta-feedback semi-supervised learning (MF-SSL), where a teacher generates pseudo labels for unlabeled driving data and is updated using
a small labeled anchor set as safety-critical feedback. In production, anchor ambiguity and
labeled-unlabeled mismatch can amplify systematic pseudo-label errors, leading to spurious
triggers. We propose a stabilized MF-SSL framework with (i) Noise-Aware Decoupling, which removes
ambiguity-prone anchors from the teacher’s supervised update path, and (ii) kinematics-gated
pseudo-labeling with a teacher conflict penalty to suppress mismatch-induced risk hallucinations on
unlabeled data while maintaining broad coverage. Extensive experiments show consistent gains as
unlabeled data scale from 1M to 1B windows, improving safety while keeping comfort stable. The
1B-trained student model is deployed to hundreds of thousands of vehicles and validated over
$10^9$ km of driving, achieving a positive-to-false activation ratio exceeding $100{:}1$ and a 35\%
improvement in accident-free driving mileage over a production rule-only baseline.
\end{abstract}

\section{INTRODUCTION}

\begin{figure}[t]
\centering
\includegraphics[width=.9\linewidth,trim=30mm 10mm 10mm 0mm,clip]{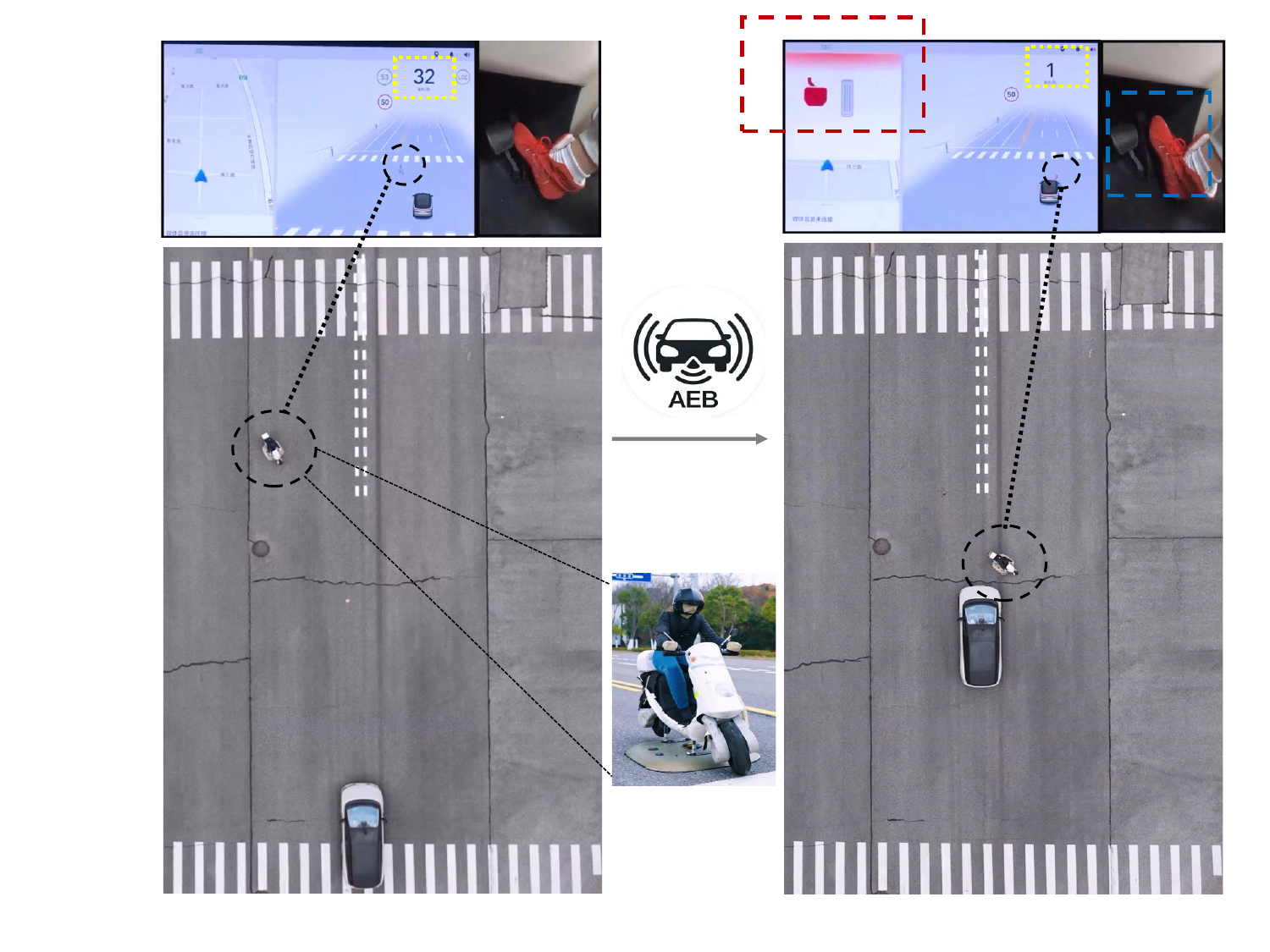}
\caption{\textbf{Test track validation before open world deployment.} One scenario is shown (left: pre-trigger; right: triggering).  The test driver does not press the brake pedal, as shown in blue dashed box, while the deployed
AEB policy autonomously triggers braking in hazardous situations. Black dashed circles highlight the test-device motorcycle; the red dashed
box marks the AEB trigger indication on the human--machine interface; and the two yellow boxes show the ego speed drop after triggering.}
\label{fig:Deployment2}
\end{figure}

Automatic Emergency Braking (AEB) detects imminent collision hazards with vehicles and vulnerable road users (VRUs) via onboard sensors and autonomously applies braking to mitigate or avoid crashes \cite{yang2022systematic, EuroNCAP2023AEB}. As shown in our test, Fig.~\ref{fig:Deployment2} illustrates a test-track triggering example. It delivers remarkable real-world safety gains, cutting rear-end crash and injury rates by 50\% and 56\% \cite{IIHS2025CrashAvoidance}, daylight pedestrian crash and injury rates by 27\% and 30\% \cite{IIHS2022PedestrianAEB}, and fatal/serious pedestrian injury odds by 20\% in unavoidable collisions \cite{Inada2025AEBPedestrian}. As a cornerstone of advanced driver-assistance systems (ADAS) and autonomous driving, AEB is now mandated for new vehicles in most major global markets \cite{NHTSA2023AEBStandard, EuroNCAP2023SafeDrivingV103_dup}.

Designing an AEB is challenging because it must
reason about ego dynamics and surrounding agents under open-world uncertainty, where handcrafted
rules based on TTC (Time-To-Collision) or distance thresholds can be brittle under sensor noise, tracking artifacts,
and long-tail interactions \cite{brannstrom2008situation, lee1976theory}. This motivates
learning-based AEB, which aims to learn trigger policies from data and generalize across diverse
driving conditions \cite{razeena2025deep, isele2018safe, chae2017autonomous}.
Despite promising results, learning-based AEB is difficult to scale in production because it
depends on safety labels that are expensive to obtain. Meanwhile, production fleets generate massive unlabeled data at essentially no marginal
annotation cost. A natural question is whether AEB can exhibit a reliable ``scaling with data''
behavior by leveraging abundant unlabeled data together with a relatively small labeled set.

We investigate semi-supervised learning (SSL) for scaling learning based AEB with massive unlabeled data. SSL has demonstrated strong data scaling behavior in vision recognition and detection,
where increasing unlabeled data can substantially improve performance when paired with limited
labels \cite{xie2020self}. In this work, we focus on meta-feedback SSL~\cite{MPL}, a teacher-student procedure in which
labeled examples serve as \emph{safety anchors}. The teacher generates pseudo labels on unlabeled
data to train the student, and is then updated based on the student's improvement on the safety
anchors, encouraging pseudo labels that benefit the student while remaining aligned with
safety-critical supervision.

Applying SSL to production AEB faces two practical obstacles. First, labeled anchors
often contain near boundary ambiguity: different annotators have different risk tolerances,
creating noisy supervision around the brake trigger decision. Second, the labeled set is typically
curated and event heavy, while the unlabeled data is dominated by nominal driving, leading to a pronounced labeled-unlabeled mismatch. Under these conditions, systematic pseudo-label errors can even grow as unlabeled data scale, undermining the stability and effectiveness of SSL.

We propose a production-oriented framework that stabilizes meta-feedback SSL, enabling reliable scaling of learning-based AEB with massive unlabeled fleet data despite label ambiguity and labeled-unlabeled mismatch. Our approach demonstrates strong effectiveness in both simulation evaluation and large-scale real-world deployment. 

Our main contributions are summarized as follows:
\begin{itemize}
\item We study meta-feedback semi-supervised learning (MF-SSL) for safety-critical AEB under production constraints. We provide two complementary analysis views: (i) a safety-generalization decomposition that highlights the role of accepted pseudo-label errors and unlabeled coverage in deployment risk, and (ii) a meta-gradient coupling view that explains how anchor ambiguity and labeled-unlabeled mismatch can amplify systematic pseudo-label errors in the teacher-student loop.

\item We propose a stabilized MF-SSL framework for AEB, combining Noise-Aware Decoupling and kinematics-gated pseudo-labeling with a teacher conflict penalty. This design suppresses high-confidence pseudo-label contradictions while maintaining high unlabeled acceptance coverage. Extensive ablations and scaling experiments show consistent gains as unlabeled data scale up to the billion-sample ($10^9$) regime.

\item To the best of our knowledge, this work is the \textbf{first learning-based AEB model deployed in mass production}. We build a full-stack data loop integrating SSL training, closed-loop evaluation, on-board deployment, and iterative data collection. Validated over $\mathbf{10^9}$ km of real-world driving across hundreds of thousands of vehicles over six months, our model achieves a positive-to-false activation ratio exceeding $\mathbf{100{:}1}$ and improves accident-free driving mileage by $\mathbf{35\%}$ over the rule-only baseline.
\end{itemize}

\section{RELATED WORK}

\subsection{Rule-based and Learning-based AEB}
AEB triggers braking to mitigate or avoid collisions~\cite{yang2022systematic, EuroNCAP2023AEB}.
Traditional AEB relies on hand-crafted rules such as TTC/safe-distance thresholds and kinematic
braking envelopes~\cite{brannstrom2008situation, lee1976theory, han2014development}. Such rules
provide strong safety priors but can be brittle under sensor noise, tracking artifacts, and
open-world variability.

Learning-based AEB uses data-driven models to better capture temporal context and multi-agent
interactions. Supervised approaches learn braking decisions or risk surrogates from annotated data,
from decision models on perception features~\cite{razeena2025deep} to spatiotemporal networks for
TTC/risk estimation~\cite{teng2024dttcnet}. Hybrid designs combine rule triggers with learned modules
to refine braking decisions while retaining safety priors~\cite{zhang2024dual}. Reinforcement
learning has also been explored in simulation with safety constraints or reward shaping
\cite{isele2018safe, chae2017autonomous}. Most prior work is evaluated on controlled datasets or simulators; scaling learning-based AEB to production fleets under sparse supervision and open-world conditions remains underexplored.

\subsection{SSL: Scaling Potential and Production Challenges}

SSL leverages unlabeled data via pseudo-labeling and consistency regularization
\cite{lee2013pseudo}, and has demonstrated strong scaling behavior in large-vision
settings, where increasing unlabeled data can substantially improve performance when coupled with a
small labeled set \cite{xie2020self}. Meta-feedback approaches such
as MPL~\cite{MPL} further use labeled data to provide a feedback signal that steers the teacher
generating pseudo labels for the unlabeled pool, which can serve as labeled \emph{safety anchors}
for AEB supervision. Beyond benchmarks, SSL has also been adopted in practical
perception pipelines where annotation is expensive, often achieving performance stronger than
supervised-only baselines by exploiting massive unlabeled data. In many real applications,
high-confidence predictions do not necessarily imply correctness under distribution shift, occlusion,
and long-tail conditions; hence a core design principle is to introduce \emph{task-specific
structure} that filters or corrects pseudo labels (e.g., geometric/physical constraints,
uncertainty-aware selection, or consistency under perturbations)
\cite{lee2024learning}. 

This line of work is attractive for production AEB because fleet data are abundant and largely
unlabeled. However, existing SSL and meta-feedback methods are mostly developed for perception or
benchmark settings, and their stability under safety-critical trigger learning and real-world
distribution shift has received limited attention. Our work builds on meta-feedback SSL and focuses on
stabilizing its scaling behavior for production AEB.

The remainder of this paper is organized as follows. Sec.~\ref{sec:safety_bound} analyzes
meta-feedback SSL for AEB scaling and motivates our stabilization design. Sec.~\ref{sec:method}
presents the proposed training framework and algorithm. Sec.~\ref{sec:exp} describes the
experimental setup and reports results on simulation evaluation, and large-scale
fleet deployment. Sec.~\ref{sec:con} concludes the paper.

\section{Analyzing Meta-Feedback SSL for AEB}
\label{sec:safety_bound}

\subsection{Preliminaries}
We model learning-based AEB as a triggering policy $f_\theta$, parameterized by $\theta$, that maps a $T$-frame history window
$x:=\mathbf{X}_{t-T+1:t}$ to a risk score $\hat p_t\in[0,1]$ at time $t$, with a binary trigger
decision $\hat y_t=\mathbb{I}[\hat p_t>\tau]\in\{0,1\}$, where $\mathbb{I}[\cdot]$ is the indicator function and $\tau$ is a predefined trigger threshold. Here $\mathbf{X}_{t-T+1:t}$ contains ego
and tracked agent states (e.g., position, velocity, and acceleration) over the past $T$ frames.

Meta-feedback SSL~\cite{MPL} maintains a teacher model $f_{\theta_T}$ and a student model
$f_{\theta_S}$. Let $P_L$ denote a small labeled anchor set $(x,\tilde y)$ and $P_U^X$ denote a
large unlabeled data distribution over inputs $x$. The anchor label $\tilde y$ may be noisy,
especially near the trigger boundary. For each unlabeled sample $x\sim P_U^X$, the teacher produces a pseudo label
$p_{\theta_T}(x)=\mathrm{softmax}(f_{\theta_T}(x))$ and an acceptance weight
$m_{\theta_T}(x)\in\{0,1\}$ indicating whether the pseudo label is used. The student learns by
matching accepted pseudo labels:
\begin{equation}
R_U^{\mathrm{PL}}(\theta_S;\theta_T)
\triangleq
\mathbb{E}_{x\sim P_U^X}\Big[
m_{\theta_T}(x)\,\mathrm{CE}\big(f_{\theta_S}(x), p_{\theta_T}(x)\big)
\Big],
\label{eq:rpl_unlabeled}
\end{equation}
where $\mathrm{CE}(\cdot, \cdot)$ denotes the standard cross-entropy loss.

A meta-feedback iteration consists of three steps. First, a one-step update is
applied to the student using unlabeled pseudo labels:
\begin{equation}
\theta_S^{+}(\theta_T)
=\theta_S-\alpha\nabla_{\theta_S}R_U^{\mathrm{PL}}(\theta_S;\theta_T),
\label{eq:safe_lookahead}
\end{equation}
where $\alpha$ is the student learning rate and $\theta_S^{+}$ depends on $\theta_T$ through the
pseudo labels and masks. Second, the look-ahead student is evaluated on labeled anchors:
\begin{equation}
\tilde J(\theta_T)
=\mathbb{E}_{(x,\tilde y)\sim P_L}\Big[\ell\big(f_{\theta_S^{+}(\theta_T)}(x),\tilde y\big)\Big],
\label{eq:safe_meta}
\end{equation}
where $\ell(\cdot, \cdot)$ is the supervised loss function evaluated on the labeled data.
Finally, the teacher is updated by differentiating the anchor loss through the look-ahead step:
\begin{equation}
\theta_T^{+}
=\theta_T-\beta\nabla_{\theta_T}\tilde J(\theta_T),
\label{eq:safe_teacher_update}
\end{equation}
where $\beta$ is the teacher learning rate. Intuitively, this encourages pseudo labels that not
only train the student on unlabeled data but also improve anchor performance.

\subsection{Two Views of Meta-Feedback SSL for AEB Scaling}
\label{subsec:two_views}

To motivate our proposed stabilizers, we first analyze how label ambiguity and distribution mismatch affect the deployment risk in MF-SSL. We present two complementary views: View I decomposes the risk to identify key error sources, while View II explains how the closed-loop training can inadvertently amplify these errors.

Deployment risk is defined as
\begin{equation}
R_D(\theta)\triangleq \mathbb{E}_{(x,y)\sim P_D}\big[\ell(f_{\theta}(x),y)\big],
\label{eq:rd_def}
\end{equation}
where $P_D$ is the deployment distribution (with $P_D^X$ denoting its input marginal), $y$ is the ground-truth trigger label, and the loss is assumed to be bounded as $0\le \ell\le B$. Although the standard cross-entropy loss used in practice is theoretically unbounded, it is practically bounded during training due to softmax saturation and gradient clipping; we adopt this bounded assumption here to provide a tractable theoretical framework.

\noindent\textbf{View I: Safety generalization decomposition.} Intuitively, the deployment risk is governed by three factors: (i) how well the student matches
teacher pseudo labels on the accepted region, (ii) how often the teacher accepts incorrect pseudo
labels (accepted-error mass), and (iii) how much of the deployment distribution is left uncovered
by the acceptance mask (one minus coverage). Define the accepted-error mass and coverage on $P_D$:
$\Gamma_D(\theta_T)=\mathbb{P}_{(x,y)\sim P_D}(m_{\theta_T}(x)=1,\hat y_T(x)\neq y)$ and
$q_D(\theta_T)=\mathbb{P}_{x\sim P_D^X}(m_{\theta_T}(x)=1)$, 
where $\hat y_T(x) = \mathbb{I}[p_{\theta_T}(x) > \tau]$ is the teacher's binary pseudo-label decision. This can be formalized as:
\begin{equation}
R_D(\theta_S)
\le
R_D^{\mathrm{PL}}(\theta_S;\theta_T)
+ B\,\Gamma_D(\theta_T)
+ B\big(1-q_D(\theta_T)\big),
\label{eq:safe_bound_simple}
\end{equation}
where
$R_D^{\mathrm{PL}}(\theta_S;\theta_T)=
\mathbb{E}_{x\sim P_D^X}[m_{\theta_T}(x)\ell(f_{\theta_S}(x),\hat y_T(x))]$.

\noindent\textit{Implication.} 
Eq.~\eqref{eq:safe_bound_simple} follows by splitting $R_D(\theta_S)$ into the accepted region and the uncovered region. The uncovered part contributes
at most $B(1-q_D)$ since $0\le \ell\le B$, and on the accepted part replacing $y$ with $\hat y_T(x)$
incurs additional loss only when the accepted pseudo label is incorrect, which occurs with
probability $\Gamma_D$, contributing at most $B\Gamma_D$.

The bound above is stated on the deployment distribution. In training, the student minimizes the
same masked pseudo-label loss but on unlabeled fleet data, i.e.,
$R_U^{\mathrm{PL}}(\theta_S;\theta_T)$, which is defined under the input marginal $P_U^X$. The
corresponding deployment quantity is $R_D^{\mathrm{PL}}(\theta_S;\theta_T)$ evaluated under $P_D^X$.
With bounded losses,
$
R_D^{\mathrm{PL}}(\theta_S;\theta_T)
\le
R_U^{\mathrm{PL}}(\theta_S;\theta_T)
+
\mathcal{D}(P_U^X,P_D^X),$
where $\mathcal{D}(P_U^X,P_D^X)$ upper bounds the expectation gap caused by the difference between
$P_U^X$ and $P_D^X$. In plain terms, the closer $P_U^X$ is to $P_D^X$, the smaller this gap. Using larger and more
diverse unlabeled fleet data helps make $P_U^X$ closer to $P_D^X$, which improves transfer to
deployment.

\noindent\textbf{View II: Meta-gradient coupling (why $\Gamma_D$ can grow).}
The bound view identifies $\Gamma_D(\theta_T)$ as a key deployment risk driver. We now show how
meta-feedback can \emph{increase} $\Gamma_D$ in production due to a closed-loop coupling between
pseudo-label learning and anchor supervision.

\emph{Closed-loop coupling.}
From Eq.~\eqref{eq:safe_meta} and the chain rule,
\begin{equation}
\nabla_{\theta_T}\tilde J(\theta_T)
=\Big(\tfrac{\partial\theta_S^{+}}{\partial\theta_T}\Big)^{\!\top}g_L^{+},
\quad
g_L^{+}:=\nabla_{\theta_S^{+}}\mathbb{E}_{P_L}\!\left[\ell(f_{\theta_S^{+}}(x),\tilde y)\right].
\label{eq:view2_chain}
\end{equation}
Using Eq.~\eqref{eq:safe_lookahead}, $\tfrac{\partial \theta_S^{+}}{\partial \theta_T}
=-\alpha\,\nabla_{\theta_T}\nabla_{\theta_S}R_U^{\mathrm{PL}}(\theta_S;\theta_T)$, hence
\begin{equation}
\nabla_{\theta_T}\tilde J(\theta_T)
=
-\alpha\Big(\nabla_{\theta_T}\nabla_{\theta_S}R_U^{\mathrm{PL}}\Big)^{\!\top}g_L^{+}.
\label{eq:view2_star}
\end{equation}
Eq.~\eqref{eq:view2_star} makes the coupling explicit: teacher updates depend on how pseudo-label
learning changes the student, and are therefore sensitive to systematic pseudo-label artifacts.

\emph{Accepted-error injection.}
Eq.~\eqref{eq:view2_star} shows that teacher updates are driven by the cross-derivative
$\nabla_{\theta_T}\nabla_{\theta_S}R_U^{\mathrm{PL}}$. When pseudo labels contain systematic errors, the
student gradient $\nabla_{\theta_S}R_U^{\mathrm{PL}}$ is biased, and this bias backpropagates to the
teacher through $\nabla_{\theta_T}\nabla_{\theta_S}R_U^{\mathrm{PL}}$. Anchor noise further
distorts the feedback signal $g_L^{+}$ in Eq.~\eqref{eq:view2_chain}, making the teacher update
sensitive to such systematic pseudo-label artifacts.

\emph{Amplification through the closed loop.}
Let $\Gamma_D(\theta_T)$ denote the probability that the teacher accepts an incorrect pseudo label
in deployment (View~I). Under the teacher update
$\theta_T^{+}=\theta_T-\beta\nabla_{\theta_T}\tilde J(\theta_T)$, a first-order expansion gives
\begin{equation}
\Gamma_D(\theta_T^{+})
\approx
\Gamma_D(\theta_T)
-\beta\,\big\langle \nabla_{\theta_T}\Gamma_D(\theta_T),\ \nabla_{\theta_T}\tilde J(\theta_T)\big\rangle.
\label{eq:view2_taylor}
\end{equation}

In production, anchor noise and labeled-unlabeled mismatch can make $\nabla_{\theta_T}\tilde J$
favor pseudo labels that reduce anchor loss on event-heavy anchors while increasing wrong
acceptance on parts of the deployment distribution, which can raise $\Gamma_D$ and be recursively
reinforced through the teacher-student loop (in practice, this manifests as spurious high-risk
predictions and seemingly inexplicable false activations under nominal driving). Simply tightening thresholds to suppress these
errors often sacrifices coverage, limiting the benefit of scaling unlabeled data; thus our design
targets both anchor noise mitigation and mismatch induced pseudo label error suppression.

\begin{figure*}[t]
\centering
\includegraphics[width=\textwidth,trim=34mm 51mm 38mm 38mm,clip]{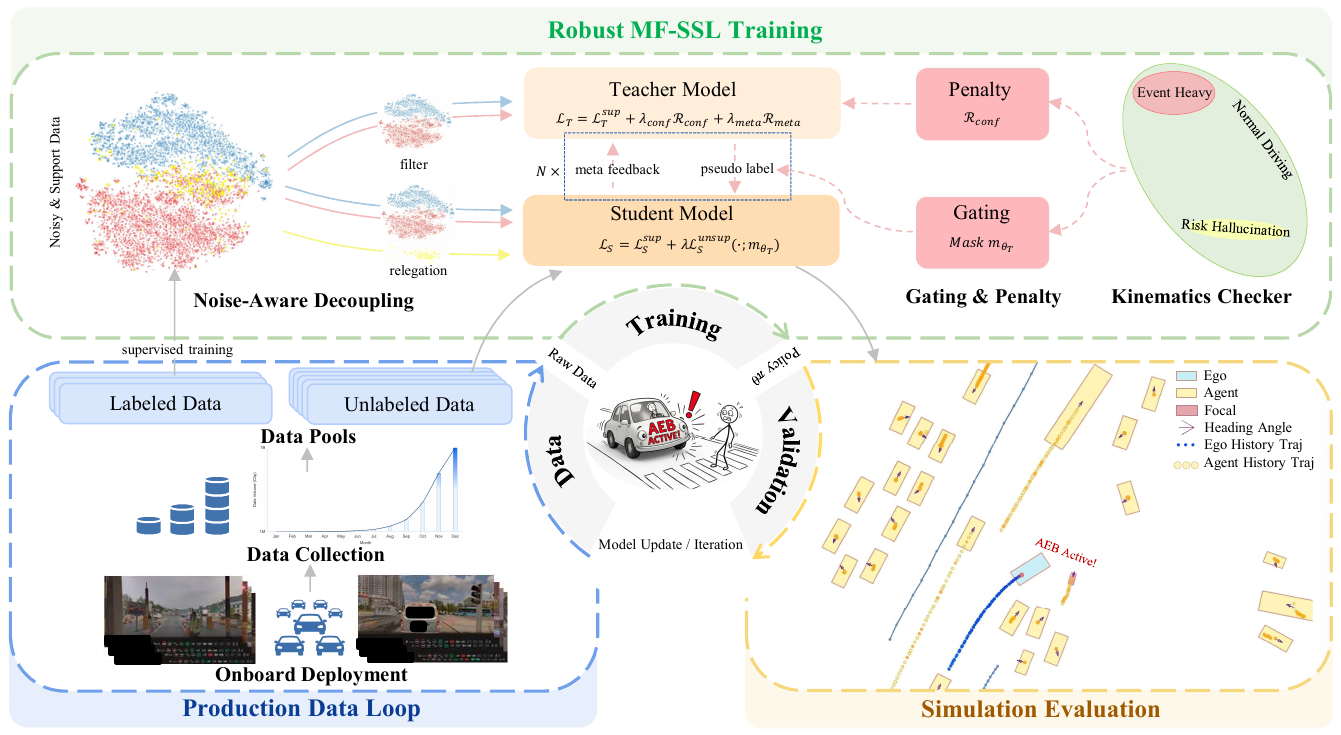}
\caption{\textbf{End-to-end workflow for production scaling.} A closed-loop production pipeline
scales learning-based AEB with massive unlabeled fleet data. \textbf{Top}: robust meta-feedback
SSL training, where Noise-Aware Decoupling reduces anchor-noise injection, and a kinematics
checker drives pseudo-label gating and a teacher conflict penalty to suppress mismatch-induced risk hallucinations. \textbf{Bottom-right}: closed-loop simulation
evaluation screens candidates before staged on-vehicle rollout. \textbf{Bottom-left}: continuous
fleet collection builds labeled anchors and a large unlabeled pool; newly collected data are fed
back for iterative updates.}
\label{fig:workflow}
\end{figure*}

\section{METHODOLOGY}
\label{sec:method}

This section presents our production-oriented framework for scaling learning-based AEB with
massive unlabeled fleet data and the end-to-end workflow used in practice
(Fig.~\ref{fig:workflow}). The workflow forms a closed data loop: fleet collection builds labeled
anchors and a large unlabeled pool; robust meta-feedback SSL trains a AEB policy; candidates are screened by closed-loop simulation before deployment, and
new data are fed back for the next iteration.

Our method comprises (i) a Transformer-based AEB architecture, (ii) Noise-Aware Decoupling to
mitigate anchor-noise injection, and (iii) a stabilized meta-feedback teacher-student SSL
procedure with kinematics-guided pseudo-label gating and a teacher conflict penalty to suppress
mismatch-induced pseudo-label errors while maintaining high unlabeled coverage. Details of the
evaluation and deployment protocol are provided in the experimental section
(Sec.~\ref{sec:exp}).

\begin{figure}[t]
\centering
\includegraphics[width=.9\textwidth,trim=56.2mm 84.9mm 65mm 36mm,clip]{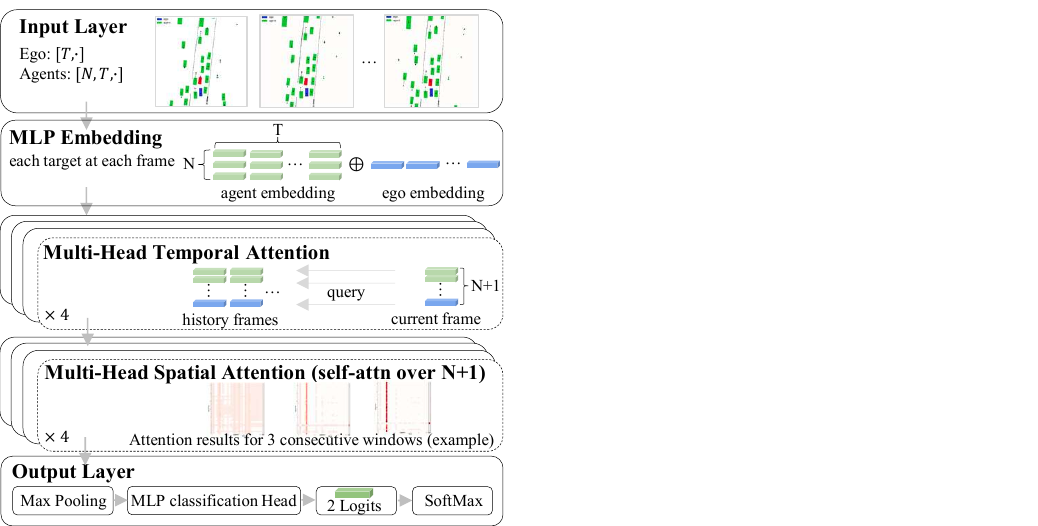}
\caption{\textbf{Model Architecture.}}
\label{fig:model_and_replay}
\end{figure}

\subsection{Model Architecture}
\label{subsec:model}
A generic Transformer-based backbone is used
(Fig.~\ref{fig:model_and_replay}), allowing the trigger boundary to be learned from data and the
training objectives rather than hand-crafted heuristics. The model takes a $T{=}60$-frame history
window $\mathbf{X}_{t-T+1:t}$ containing ego and tracked agent states (position, velocity,
acceleration, and yaw) in the ego-centric coordinate frame, and outputs the trigger probability at
time $t$. A shared MLP encoder embeds per-agent per-frame states into a $d{=}64$ dimensional feature
space. Temporal multi-head self-attention (4 heads) aggregates each agent's $T$-step history into a
per-agent embedding. Spatial multi-head self-attention (4 heads) then models interactions among
the ego and all tracked agents, followed by scene pooling to obtain a global context vector. A
classification head outputs two logits $\mathbf{z}_t$, which are converted to class probabilities
via a softmax:
\begin{equation}
\mathbf{z}_t=f_{\theta}(\mathbf{X}_{t-T+1:t})\in\mathbb{R}^2,
\qquad
{\mathbf{p}}_t=\mathrm{softmax}(\mathbf{z}_t).
\end{equation}

\subsection{Noise-Aware Decoupling (NAD)}
\label{subsec:e2u}
Labeled anchors near the trigger boundary are often ambiguous, injecting bias into teacher updates. As shown in Fig.~\ref{fig:workflow} (top-left), misclassified samples (yellow) concentrate near the boundary between positive (red) and negative (blue) clusters. These samples suffer from near-trigger ambiguity and labeling artifacts, making them unreliable as safety anchors.

To mitigate this, we train a supervised warm-up model $f_{\theta^{(0)}}$ on the labeled set $\mathcal{D}_L$ and collect its misclassified samples:
\begin{equation}
\mathcal{D}_{\mathrm{err}}
=\{(x,y)\in\mathcal{D}_L:\arg\max f_{\theta^{(0)}}(x)\neq y\}.
\end{equation}
We use misclassification rather than confidence-based filtering because modern networks are often poorly calibrated, making misclassification a more robust indicator of inherent label-feature contradiction. 

Instead of discarding these ambiguous samples, we relegate their inputs to the unlabeled pool: $\mathcal{D}_L'=\mathcal{D}_L\setminus \mathcal{D}_{\mathrm{err}}$ and $\mathcal{D}_U'=\mathcal{D}_U\cup\{x:(x,y)\in\mathcal{D}_{\mathrm{err}}\}$. We apply this decoupling within a short temporal window around the annotated trigger onset, where label ambiguity is most pronounced.

\subsection{Stabilized Meta-Feedback SSL with Kinematics Gating and Conflict Penalties}
\label{subsec:ssl}
Motivated by the analysis in Sec.~\ref{sec:safety_bound}, we stabilize meta-feedback SSL as follows.
\paragraph{Teacher-student setup.}
Training uses a teacher $f_{\theta_T}$ and a student $f_{\theta_S}$ on
$(\mathcal{D}_L',\mathcal{D}_U')$. The student learns from labeled anchors in $\mathcal{D}_L'$, and
from teacher pseudo labels on $\mathcal{D}_U'$ filtered by a kinematics gate. The teacher is
trained on anchors and is additionally regularized by a conflict penalty and a meta-feedback term.
Alg.~\ref{alg:s2aeb_full} stage 2 summarizes the full procedure.

\paragraph{Student objectives.}
For labeled anchors $(x_L,y_L)\sim\mathcal{D}_L'$, we use the standard cross-entropy loss
$\mathrm{CE}(\cdot,\cdot)$ between predicted logits and the binary label:
\begin{equation}
\mathcal{L}_S^{\mathrm{sup}}(\theta_S)
=
\mathbb{E}_{(x_L,y_L)\sim\mathcal{D}_L'}\big[\mathrm{CE}(f_{\theta_S}(x_L),y_L)\big].
\end{equation}
For unlabeled samples $x_U\sim\mathcal{D}_U'$, the teacher provides a pseudo label
$\hat y_U=\mathrm{softmax}(f_{\theta_T}(x_U))$ and an acceptance mask
$m_{\theta_T}(x_U)$ (defined below).
The masked pseudo-label loss is
\begin{equation}
\label{eq:student_unsup}
\mathcal{L}_S^{\mathrm{unsup}}(\theta_S;\theta_T)
=
\mathbb{E}_{x_U\sim\mathcal{D}_U'}\Big[
m_{\theta_T}(x_U)\cdot \mathrm{CE}(f_{\theta_S}(x_U),\hat y_U)
\Big].
\end{equation}
The total student objective is
\begin{equation}
\mathcal{L}_S(\theta_S)
=
\mathcal{L}_S^{\mathrm{sup}}(\theta_S)
+\lambda\,\mathcal{L}_S^{\mathrm{unsup}}(\theta_S;\theta_T).
\label{eq:student_total_full}
\end{equation}

\paragraph{Kinematics checker and pseudo-label gating.}
Labeled anchors are scarce and dominated by hazardous events, whereas the unlabeled pool is
$3$--$4$ orders of magnitude larger and mostly consists of nominal safe driving, leading to a
pronounced labeled-unlabeled mismatch (Fig.~\ref{fig:workflow}, top-right). This can cause the
teacher to over-confidence in obviously safe scenes (risk hallucination). We therefore use a
lightweight TTC-based checker that outputs $\mathbb{I}_{\mathrm{safe}}(x_U)$, where large TTC
indicates clearly safe kinematics. Let the teacher output class probabilities $p_T(x_U)$ and
denote the trigger probability by $p_T^{\mathrm{risk}}(x_U)$. Given a confidence threshold $\tau$,
the high-confidence conflict of risk hallucination is
\begin{equation}
\label{eq:conflict_defs_full}
\mathbb{I}_{\mathrm{hall}}(x_U)
=
\mathbb{I}\!\left[p_T^{\mathrm{risk}}(x_U)>\tau\right]\cdot \mathbb{I}_{\mathrm{safe}}(x_U).
\end{equation}
The acceptance mask filters pseudo labels that contradict the checker:
\begin{equation}
\label{eq:mask_full}
m_{\theta_T}(x_U)=1-\mathbb{I}_{\mathrm{hall}}(x_U).
\end{equation}

\paragraph{Teacher objectives.}
The teacher is trained with three terms. First, supervised learning on anchors:
\begin{equation}
\mathcal{L}_T^{\mathrm{sup}}(\theta_T)
=
\mathbb{E}_{(x_L,y_L)\sim\mathcal{D}_L'}\big[
\mathrm{CE}(f_{\theta_T}(x_L),y_L)
\big].
\end{equation}
Second, a conflict penalty regularizes the teacher by discouraging high-confidence predictions
that contradict the checker:
\begin{equation}
\mathcal{R}_{\mathrm{conf}}(\theta_T)
=
\mathbb{E}_{x_U\sim\mathcal{D}_U'}\Big[
\mathbb{I}_{\mathrm{hall}}(x_U)\,p_T^{\mathrm{risk}}(x_U)
\Big].
\label{eq:teacher_conf_penalty_full}
\end{equation}
Third, meta-feedback updates the teacher using anchor performance after a one-step look-ahead
student update on unlabeled data:
\begin{equation}
\theta_S^{+}(\theta_T)
=
\theta_S-\eta_S \nabla_{\theta_S}\mathcal{L}_S^{\mathrm{unsup}}(\theta_S;\theta_T),
\label{eq:lookahead_hard}
\end{equation}
\begin{equation}
\mathcal{R}_{\mathrm{meta}}(\theta_T)
:=
\mathcal{L}_S^{\mathrm{sup}}\!\big(\theta_S^{+}(\theta_T)\big).
\label{eq:teacher_meta_hard}
\end{equation}
The total teacher objective is
\begin{equation}
\mathcal{L}_T(\theta_T)
=
\mathcal{L}_T^{\mathrm{sup}}(\theta_T)
+\lambda_{\mathrm{conf}}\mathcal{R}_{\mathrm{conf}}(\theta_T)
+\lambda_{\mathrm{meta}}\mathcal{R}_{\mathrm{meta}}(\theta_T).
\label{eq:teacher_total_full}
\end{equation}

\begin{algorithm}[t]
\caption{Robust Meta-Feedback Semi-Supervised Training Framework}
\label{alg:s2aeb_full}
\begin{algorithmic}[1]
\REQUIRE Labeled $\mathcal{D}_L$, unlabeled $\mathcal{D}_U$, threshold $\tau$, weights $\lambda,\lambda_{\mathrm{conf}},\lambda_{\mathrm{meta}}$
\STATE \textbf{Stage 1: Noise-Aware Decoupling} \quad $(\mathcal{D}_L',\mathcal{D}_U')\leftarrow \textsc{NAD}(\mathcal{D}_L,\mathcal{D}_U)$
\STATE \textbf{Stage 2: Teacher-Student SSL} \quad Initialize $\theta_S,\theta_T$
\FOR{each iteration}
    \STATE Sample minibatch $\{(x_L,y_L)\}\sim\mathcal{D}_L'$ and $\{x_U\}\sim\mathcal{D}_U'$
    \STATE Teacher forward: $\hat y_U=\mathrm{softmax}(f_{\theta_T}(x_U))$
    \STATE Compute $\mathbb{I}_{\mathrm{safe}}(x_U)$ via kinematic check
    \STATE Compute $\mathbb{I}_{\mathrm{hall}}(x_U)$ by Eq.~\eqref{eq:conflict_defs_full}
    \STATE Mask $m_{\theta_T}(x_U)$ by Eq.~\eqref{eq:mask_full}
    \STATE \textbf{Student update:} minimize Eq.~\eqref{eq:student_total_full} to update $\theta_S$
    \STATE \textbf{Look-ahead:} compute $\theta_S^{+}$ by Eq.~\eqref{eq:lookahead_hard}
    \STATE \textbf{Teacher update:} minimize Eq.~\eqref{eq:teacher_total_full} (supervised + conflict penalty + meta objective)
\ENDFOR
\end{algorithmic}
\end{algorithm}

\section{EXPERIMENTS}
\label{sec:exp}

\subsection{Implement Details}
\label{subsec:exp_setup}

\paragraph{Training details.}
All models are implemented in PyTorch and optimized with AdamW using a cosine learning-rate
schedule. Batch size is 16384, learning rate is $1\times 10^{-4}$, and weight decay is 0.01, loss
weights are set to 1 (i.e., $\lambda=\lambda_{\mathrm{conf}}=\lambda_{\mathrm{meta}}=1$), and
we observe low sensitivity to these hyperparameters.

\paragraph{Datasets.}
\textbf{Labeled anchors.}
The labeled dataset $\mathcal{D}_L$ contains 10k event segments, resulting in $\sim$1M
window-level training samples with a positive (trigger) rate of 20\%.
Noise-Aware Decoupling produces a cleaned anchor set $\mathcal{D}_L'$ by
identifying $\sim$70k hard-to-fit labeled windows and removing them from the teacher’s supervised
update path. 
\textbf{Unlabeled pools.}
We construct four unlabeled datasets at different scales from raw fleet driving logs,
containing [1M, 10M, 100M, 1B] window-level samples, respectively.
The unlabeled data are collected without manual curation, follow
the natural fleet distribution dominated by nominal driving with only a small fraction of
hazardous events.
\textbf{Simulation evaluation sets.}
We use two held-out scenario sets: (i) a safety set of accident scenarios with
$N_{\mathrm{safe}}=2000$, and (ii) a comfort set of safe but potentially misleading scenarios that
may induce unnecessary AEB triggers (false activations) with $N_{\mathrm{comf}}=6000$. Each
scenario lasts 15\,s at 20\,Hz. Both sets are manually curated to cover diverse ego maneuvers
(straight, turning, U-turns) and target types (vehicles and VRUs) with behaviors including
stationary, crossing (with abrupt stop), cut-ins, and wrong-way encounters. All evaluation data
are excluded from training based on unique timestamps to prevent leakage.

\paragraph{Kinematics checker and gating.}
We apply a lightweight TTC-based checker to each tracked target using the most recent frame. It
flags clearly safe cases with $\mathbb{I}_{\mathrm{safe}}(x)=\mathbb{I}[\mathrm{TTC}(x)>10]$.
High-confidence conflicts use $\tau=0.9$ (Eq.~\eqref{eq:conflict_defs_full}), and the
kinematics-gated mask follows Eq.~\eqref{eq:mask_full}. The checker is implemented as vectorized
GPU tensor operations with negligible overhead.

\paragraph{Deployment.}
Model is deployed via over-the-air (OTA) updates to both test vehicles and a production fleet of hundreds of thousands of vehicles.
On-vehicle inference is accelerated with TensorRT on NVIDIA DRIVE Orin and Thor and runs at
20\,Hz using a timer-driven pipeline.

\subsection{Evaluation Metric}
\label{subsec:exp_protocol}

\paragraph{Simulation.}
We build a closed-loop log-replay evaluator (similar to nuPlan) with a data-driven ego braking
dynamics model. The ego follows logs before triggering and is simulated after
triggering, while other participants are replayed non-reactively. This enables
repeatable evaluation of collision mitigation on the original accident logs, as well as
false-activation severity under nominal driving. Let $v_0^{(i)}$ denote the ego speed at the first trigger frame in scenario $i$,
$v_{\mathrm{col}}^{(i)}$ the ego speed at the collision time if a collision occurs,
and $v_{\min}^{(i)}$ the minimum ego speed after triggering. Here $\mathbb{I}[\mathrm{col}]$
indicates whether a collision occurs in scenario $i$, and $\mathbb{I}[\mathrm{trig}]$ indicates
whether an AEB trigger occurs in scenario $i$. We report collision-mitigation (safety) $S_{\mathrm{safe}}$
and false-activation severity (comfort) $S_{\mathrm{comf}}$:
\begin{equation}
S_{\mathrm{safe}} \triangleq \frac{100}{N_{\mathrm{safe}}}\sum_{i=1}^{N_{\mathrm{safe}}}
\Big(1-\mathbb{I}[\mathrm{col}]\cdot
\big(1-\mathrm{clip}\!\big(\tfrac{v_0^{(i)}-v_{\mathrm{col}}^{(i)}}{v_0^{(i)}+\epsilon},0,1\big)\big)\Big),
\label{eq:ssafe}
\end{equation}
\begin{equation}
S_{\mathrm{comf}} \triangleq \frac{100}{N_{\mathrm{comf}}}\sum_{i=1}^{N_{\mathrm{comf}}}
\Big(1-\mathbb{I}[\mathrm{trig}]\cdot \mathrm{clip}\!\big(\tfrac{v_0^{(i)}-v_{\min}^{(i)}}{v_0^{(i)}+\epsilon},0,1\big)\Big).
\label{eq:scomf}
\end{equation}

\paragraph{Deployment.}
During the initial stage (the first 10M\,km), the model runs in shadow mode: trigger decisions are logged with full context but do not actuate braking. As a safety gate, all logged trigger events are manually reviewed and independently double-checked by an AEB expert operations team to determine whether each activation is a true positive or a false trigger. We compute the \textit{positive-to-false activation ratio} on this
initial 10M\,km:
\begin{equation}
\mathrm{PFR} \triangleq
\frac{\sum_{e\in\mathcal{E}_{10\mathrm{M}}}\mathbb{I}[z_e=1]}
{\max(\sum_{e\in\mathcal{E}_{10\mathrm{M}}}\mathbb{I}[z_e=0], 1)},
\end{equation}
where $\mathcal{E}_{10\mathrm{M}}$ denotes the set of reviewed trigger events in the first 10M\,km,
and $z_e\in\{0,1\}$ indicates whether event $e$ is a valid (positive) activation. Over longer
horizons with brake actuation enabled, we evaluate fleet-wide safety gains using
\textit{accident-free driving mileage}:
\begin{equation}
\mathrm{AFM} \triangleq \frac{M}{\max(N_{\mathrm{acc}},1)},
\label{eq:afm}
\end{equation}
$M$ is the total driven mileage, $N_{\mathrm{acc}}$ is the number of accidents aggregated
from user-reported feedback and onboard vehicle-body impact sensors (due to compliance
review, we report relative improvements rather than absolute counts).

\begin{table}[!b]
\centering
\caption{\textbf{Simulation results (100M unlabeled).} Higher is better.
}
\label{tab:sim_results}
\setlength{\tabcolsep}{4pt}
\renewcommand{\arraystretch}{1.06}
\begin{tabularx}{\columnwidth}{Xcc}
\toprule
Method & $S_{\mathrm{safe}}\uparrow$ & $S_{\mathrm{comf}}\uparrow$ \\
\midrule
Rule-based & 47.51 & 96.23 \\
Supervised (no SSL) & 47.90 & 92.70 \\
Self-training (PL)~\cite{lee2013pseudo} & 58.77 & 89.23 \\
Noisy Student~\cite{xie2020self} & 62.39 & 85.73 \\
Vanilla MF-SSL~\cite{MPL} & 62.43 & 81.72 \\
Ours & \textbf{67.80} & \textbf{97.67} \\
\bottomrule
\end{tabularx}
\end{table}

\subsection{Simulation Results}
\label{subsec:exp_sim_results}
Table~\ref{tab:sim_results} reports simulation results under a unified setting where all SSL
methods use 100M unlabeled windows and share the same model and optimization settings. The
rule-based baseline is our production-validated AEB logic and provides a conservative reference.
The supervised-only model is trained on event-heavy anchors and shows limited generalization to
the comfort set dominated by safe scenarios, resulting in more unnecessary triggers and a lower
$S_{\mathrm{comf}}$.

Among the SSL baselines, self-training (PL) improves collision mitigation, and Noisy Student achieves a
larger gain, but both incur a clear drop in $S_{\mathrm{comf}}$. Vanilla MF-SSL reduces
$S_{\mathrm{comf}}$ even further, likely because the meta-feedback loop can amplify training
signals and reinforce systematic pseudo-label errors in the presence of anchor noise and
labeled-unlabeled distribution shift. Our
method improves $S_{\mathrm{safe}}$ from 47.90 to 67.80 while increasing $S_{\mathrm{comf}}$ from 92.70
to 97.67, achieving the best overall balance between safety and comfort.

\begin{table}[b]
\centering
\caption{\textbf{Ablation of robust MF-SSL components on 100M unlabeled data.}
Higher is better for $S_{\mathrm{safe}}$, $S_{\mathrm{comf}}$, and $q_U$; lower is better for
$r_{\mathrm{conf}}$. We report $q_U$ as percentage (\%) and $r_{\mathrm{conf}}$ in per-mille
($\mathrm{permille}$).}
\label{tab:ablation_mfssl}
\setlength{\tabcolsep}{4.0pt}
\renewcommand{\arraystretch}{1.06}
\begin{tabularx}{\columnwidth}{Xcccc}
\toprule
Method & $S_{\mathrm{safe}}\uparrow$ & $S_{\mathrm{comf}}\uparrow$ & $q_U\uparrow$ & $r_{\mathrm{conf}}\downarrow$ \\
\midrule
Supervised & 47.90 & 92.70 & -- & 0.06 \\
Vanilla MF-SSL & 62.43 & 81.72 & 98.12 & 0.22 \\
+NAD (w/o relegation) & 64.03 & 93.85 & 97.97 & 0.09 \\
+NAD & 64.72 & 95.85 & 98.08 & 0.09 \\
+Gating & 63.30 & 96.38 & 98.71 & 0.09 \\
+ConfPen & 64.36 & 96.69 & 98.84 & 0.07 \\
+Gating+ConfPen & 66.71 & 97.16 & 98.93 & 0.05 \\
Full method & \textbf{67.80} & \textbf{97.67} & \textbf{99.21} & \textbf{0.01} \\
\bottomrule
\end{tabularx}
\end{table}

\begin{table}[b]
\centering
\caption{\textbf{Effect of meta-feedback (MF) on 100M unlabeled data.}
Higher is better for $S_{\mathrm{safe}}$ and $S_{\mathrm{comf}}$.}
\label{tab:ablation_mf}
\setlength{\tabcolsep}{6pt}
\renewcommand{\arraystretch}{1.06}
\begin{tabularx}{\columnwidth}{Xcc}
\toprule
Method & $S_{\mathrm{safe}}\uparrow$ & $S_{\mathrm{comf}}\uparrow$ \\
\midrule
Full method (w/o MF) & 65.91 & 93.83 \\
Full method (w/ MF)  & \textbf{67.80} & \textbf{97.67} \\
\bottomrule
\end{tabularx}
\end{table}

\subsection{Ablation Studies}
\label{subsec:ablation}
We analyze the impact of four components in our robust MF-SSL framework: (1) Noise-Aware Decoupling;
(2) kinematics-gated pseudo-labeling (Gating); and (3) the teacher conflict penalty (ConfPen);  (4) meta-feedback~\cite{MPL}. For NAD, we further ablate a variant \emph{w/o relegation} that removes $\mathcal{D}_{\mathrm{err}}$
from $\mathcal{D}_L$ but does not add it to the unlabeled pool. All
ablations use 100M unlabeled windows and the evaluation protocol in Sec.~\ref{subsec:exp_protocol}.
We report $S_{\mathrm{safe}}$ and $S_{\mathrm{comf}}$ from simulation evaluation
(Eq.~\eqref{eq:ssafe}--\eqref{eq:scomf}), together with unlabeled acceptance coverage
$q_U=\mathbb{E}_{x_U\sim \mathcal{D}_U'}[m_{\theta_T}(x_U)]$ (Eq.~\eqref{eq:mask_full}) and the
conflict rate
$r_{\mathrm{conf}}=\mathbb{E}_{x_U\sim \mathcal{D}_U'}[\mathbb{I}_{\mathrm{hall}}(x_U)]$
(Eq.~\eqref{eq:conflict_defs_full}) on $\mathcal{D}_U'$. In practice, $q_U$ and $r_{\mathrm{conf}}$
serve as deployment-facing proxies for unlabeled coverage and accepted pseudo-label errors in our
analysis (Sec.~\ref{sec:safety_bound}).

\noindent\textbf{Impact of NAD.}
As shown in Table~\ref{tab:ablation_mfssl}, \textbf{+NAD} substantially improves both
$S_{\mathrm{safe}}$ and $S_{\mathrm{comf}}$ over \textbf{Vanilla MF-SSL}, while keeping $q_U$ high and
reducing $r_{\mathrm{conf}}$. Moreover, \textbf{+NAD (w/o relegation)} performs worse than
\textbf{+NAD}, indicating that discarding hard-to-fit anchors can waste useful data, whereas
relegating them to the unlabeled pool allows SSL to further exploit them and stabilizes the
meta-feedback loop.

\noindent\textbf{Impact of kinematics gating (Gating).}
Comparing \textbf{Vanilla MF-SSL} and \textbf{+Gating}, Gating increases $S_{\mathrm{safe}}$ and reduces
$r_{\mathrm{conf}}$ by filtering accepted pseudo labels that contradict conservative kinematic
checks, while maintaining high acceptance coverage $q_U$.

\noindent\textbf{Impact of teacher conflict penalty (ConfPen).}
Comparing \textbf{Vanilla MF-SSL} and \textbf{+ConfPen}, the conflict penalty reduces
$r_{\mathrm{conf}}$ and improves $S_{\mathrm{comf}}$ by suppressing over-confident risk hallucinations
(i.e., high-confidence teacher predictions that contradict the checker). Combining
\textbf{+Gating+ConfPen} yields larger gains by filtering inconsistent pseudo labels for the
student and correcting them at the source.

\noindent\textbf{Effect of meta-feedback (MF).}
Table~\ref{tab:ablation_mf} compares the full method with and without meta-feedback. Removing MF
reduces both $S_{\mathrm{safe}}$ and $S_{\mathrm{comf}}$, indicating that anchor-based teacher updates
help steer pseudo labeling toward safety-critical supervision.

\begin{figure}[t]
\centering
\includegraphics[width=\linewidth]{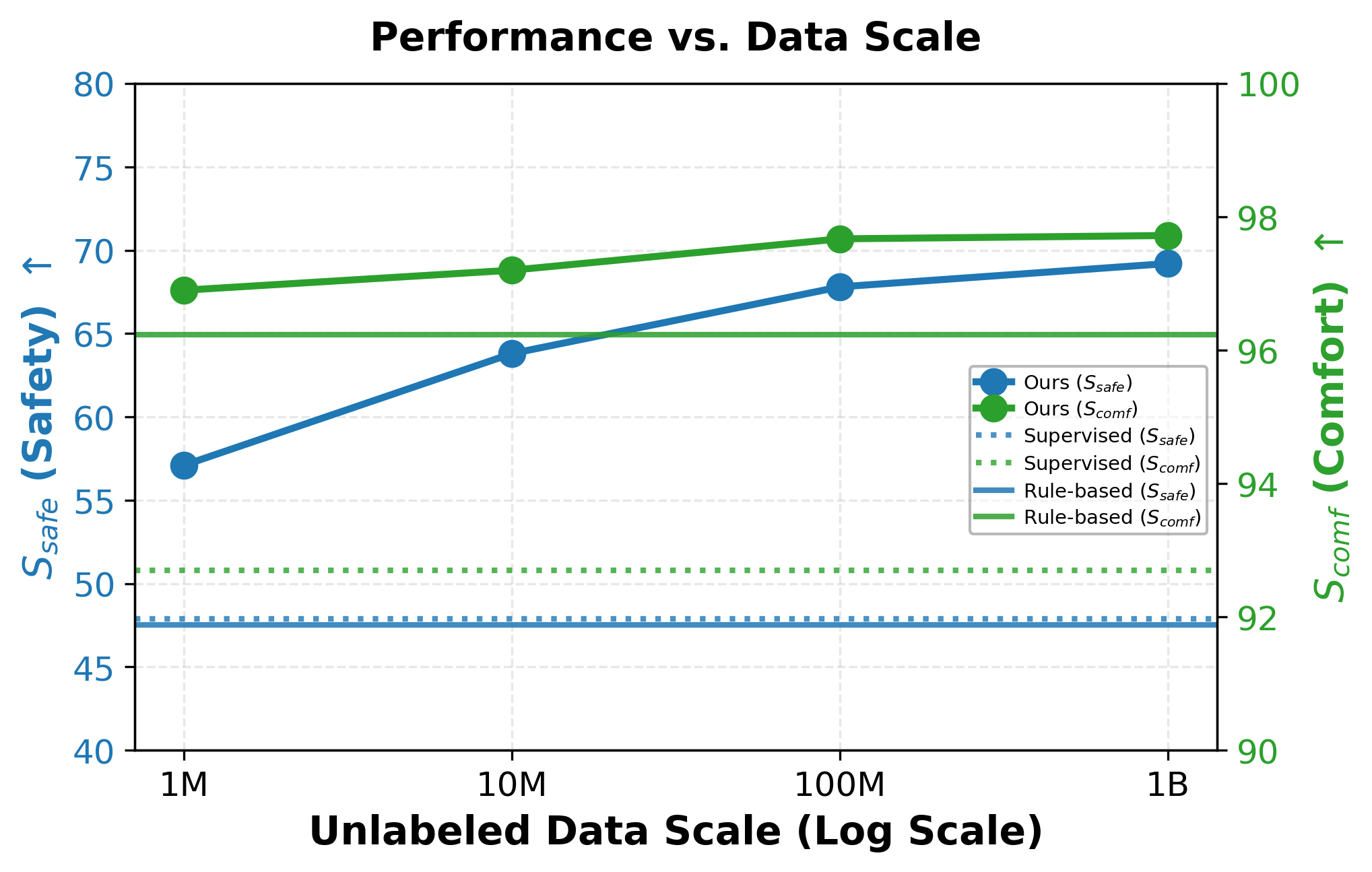}
\caption{\textbf{Scaling with massive unlabeled data.}
As the unlabeled scale increases from 1M to 1B (log scale), our method improves the
collision-mitigation score $S_{\mathrm{safe}}$ (Eq.~\eqref{eq:ssafe}) while keeping the
false-activation score $S_{\mathrm{comf}}$ (Eq.~\eqref{eq:scomf}) high and stable. The rule-based method
serves as a baseline reference.The supervised model does not leverage unlabeled
data and thus remains unchanged across data scales.}
\label{fig:scaling_curve}
\end{figure}

\begin{figure*}[t]
\centering
\includegraphics[width=\textwidth,trim=0mm 101mm 0mm 10mm,clip]{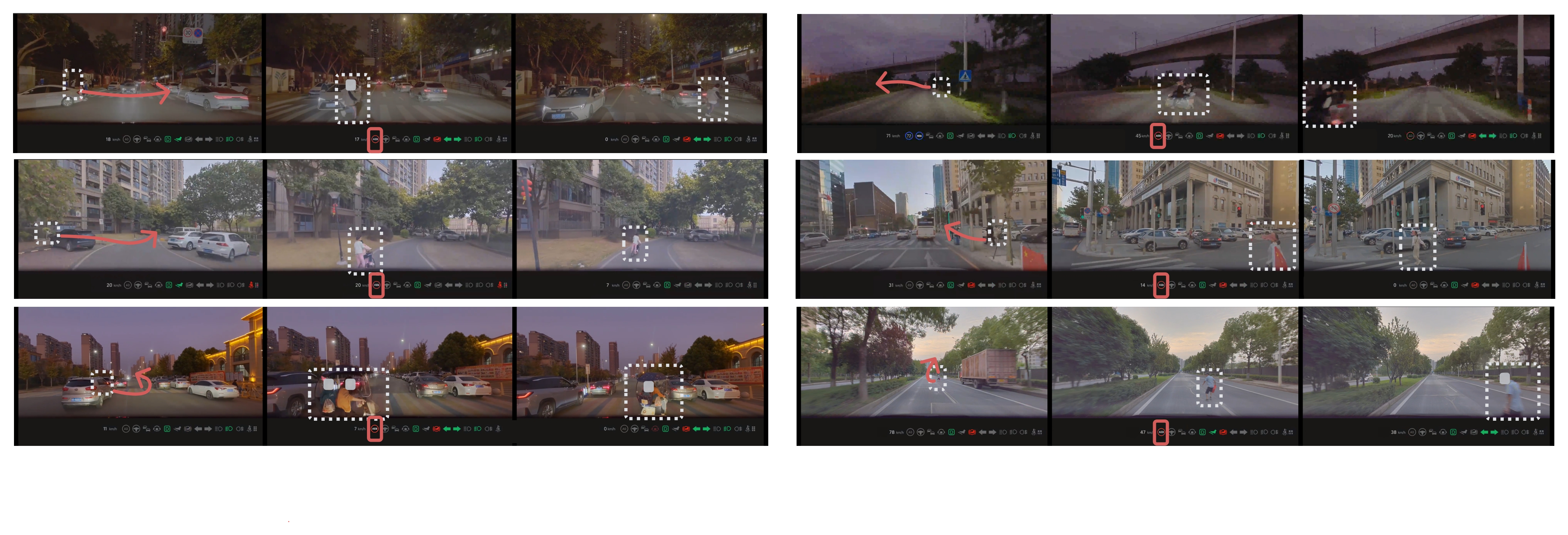}
\caption{\textbf{Challenge cases from large-scale deployment.}
Each row shows a triggering event
collected from deployment (left: pre-trigger; middle: trigger; right: post-trigger).
White dashed boxes highlight the relevant targets, red arrows in the left panel denote the
targets' future motion to help readers identify the impending collision risk, and red boxes indicate the AEB trigger signal shown on the human-machine interface. The model reacts quickly to emerging hazards and triggers braking immediately only in truly
risky situations, mitigating or avoiding the potential accident/near-collision outcome, while
maintaining comfort in the vast majority of driving.}
\label{fig:deploy_cases}
\end{figure*}

\subsection{Scaling with Massive Unlabeled Data}
\label{subsec:exp_scaling}

The unlabeled-data scaling study fixes the labeled anchors $\mathcal{D}_L'$ and the optimization
budget to 200k training steps with identical batch size and hyperparameters, and only varies the
size of the unlabeled pool. Fig.~\ref{fig:scaling_curve} shows consistent gains as unlabeled data
increase from 1M to 1B windows: $S_{\mathrm{safe}}$ improves steadily, while $S_{\mathrm{comf}}$
remains high and stable. These results suggest that the proposed SSL pipeline can effectively
exploit massive unlabeled data to expand AEB capability, substantially reducing reliance on
additional safety labeling.

\subsection{Deployment Results}
\label{subsec:fleet_deploy}

We deploy the 1B-trained student model to a production fleet of hundreds of thousands of vehicles
via OTA updates and run it for six months, accumulating over $10^9$ km of real-world driving.
The deployed model generalizes well across diverse traffic, lighting, and weather conditions
(Fig.~\ref{fig:deploy_cases}), triggering in safety-relevant situations while maintaining a low
false-activation level. During the initial 10M\,km in monitoring mode (no brake actuation), all
logged trigger events are manually reviewed, yielding a positive-to-false activation ratio (PFR)
exceeding $>100{:}1$. Over longer horizons with brake actuation enabled, we measure fleet-wide safety
using accident-free driving mileage (AFM) and observe a 35\% improvement over a production
rule-only baseline (we report relative improvements due to compliance review).
Table~\ref{tab:deploy_results} summarizes the deployment results. Fig.~\ref{fig:deploy_cases} shows representative challenge cases collected from deployment.

\begin{table}[t]
\centering
\caption{\textbf{Fleet deployment results.}
PFR: positive-to-false activation ratio, AFM: accident-free driving mileage.
}
\label{tab:deploy_results}
\setlength{\tabcolsep}{4pt}
\renewcommand{\arraystretch}{1.05}
\begin{tabularx}{\columnwidth}{Xc|cc}
\toprule
Method & Shadow Mode (first 10M km) & \multicolumn{2}{c}{Actuation Enabled} \\
\cmidrule(lr){2-2}\cmidrule(lr){3-4}
 & PFR $\uparrow$ & Mileage (km) & AFM $\uparrow$ \\
\midrule
1B-Model & $>100{:}1$ & $>10^9$ & 35\% \\
\bottomrule
\end{tabularx}
\end{table}

\section{Conclusion}
\label{sec:con}
This study shows that learning-based AEB can scale reliably with massive unlabeled fleet data when the
meta-feedback SSL loop is stabilized for production anchor ambiguity and labeled-unlabeled
mismatch. Noise-Aware Decoupling reduces ambiguous anchor influence, and kinematics-gated
pseudo-labeling with a teacher conflict penalty suppresses contradictions while maintaining broad
unlabeled coverage. These stabilizers yield consistent safety gains with stable comfort as
unlabeled data scale to the billion-sample regime, and transfer to large-scale fleet deployment
through an end-to-end industrial data loop. Future work will further study scaling and extend this approach to multimodal AEB.

\bibliographystyle{IEEEtran}
\bibliography{IEEEabrv,references}

@techreport{IIHS2025CrashAvoidance,
  title        = {Real-world benefits of crash avoidance technology},
  author       = {{Insurance Institute for Highway Safety - Highway Loss Data Institute}},
  shortauthor  = {IIHS-HLDI},
  year         = {2025},
  type         = {Technical Report},
  institution  = {Insurance Institute for Highway Safety - Highway Loss Data Institute},
  address      = {Arlington},
  note         = {Accessed: March 2, 2026}
}

@online{IIHS2022PedestrianAEB,
  title        = {Pedestrian crash avoidance systems cut crashes -- but not in the dark},
  author       = {{Insurance Institute for Highway Safety}},
  shortauthor  = {IIHS},
  year         = {2022},
  note         = {Accessed: March 2, 2026}
}

@article{Inada2025AEBPedestrian,
  title        = {Association between automatic emergency braking and pedestrian and cyclist injury severity in Japan},
  author       = {Inada, H and Ichikawa, M},
  journal      = {Accident Analysis \& Prevention},
  volume       = {198},
  pages        = {107562},
  year         = {2025},
  publisher    = {Elsevier}
}

@techreport{NHTSA2023AEBStandard,
  title        = {Federal Motor Vehicle Safety Standard for Automatic Emergency Braking (AEB) and Pedestrian AEB (PAEB)},
  author       = {{National Highway Traffic Safety Administration}},
  shortauthor  = {NHTSA},
  year         = {2023},
  type         = {Federal Motor Vehicle Safety Standard},
  institution  = {U.S. Department of Transportation},
  address      = {Washington},
  note         = {Accessed: March 2, 2026}
}

@techreport{EuroNCAP2023SafeDrivingV103_dup,
  title        = {Assessment Protocol: Safe Driving Vehicle Assistance v10.3},
  author       = {{European New Car Assessment Programme}},
  shortauthor  = {Euro NCAP},
  year         = {2023},
  type         = {Assessment Protocol},
  institution  = {European New Car Assessment Programme Secretariat},
  address      = {Brussels},
  note         = {Accessed: March 2, 2026}
}

@InProceedings{MPL,
    author    = {Pham, Hieu and Dai, Zihang and Xie, Qizhe and Le, Quoc V.},
    title     = {Meta Pseudo Labels},
    booktitle = {CVPR},
    month     = {June},
    year      = {2021},
    pages     = {11557-11568}
}

@article{yang2022systematic,
  title={A systematic review of autonomous emergency braking system: impact factor, technology, and performance evaluation},
  author={Yang, Lan and Yang, Yipeng and Wu, Guoyuan and Zhao, Xiangmo and Fang, Shan and Liao, Xishun and Wang, Runmin and Zhang, Mengxiao},
  journal={J. Adv. Transp.},
  volume={2022},
  number={1},
  pages={1188089},
  year={2022},
  publisher={Wiley Online Library}
}

@article{teng2024dttcnet,
  title={DTTCNet: Time-to-Collision Estimation With Autonomous Emergency Braking Using Multi-Scale Transformer Network},
  author={Teng, Xiaoqiang and Xu, Shibiao and Guo, Deke and Guo, Yulan and Meng, Weiliang and Zhang, Xiaopeng},
  journal={IEEE TMC},
  year={2024},
  publisher={IEEE}
}

@techreport{EuroNCAP2023AEB,
  title        = {{Euro NCAP} AEB Car-to-Car Test Protocol, Version 4.2},
  author       = {{European New Car Assessment Programme}},
  year         = {2023},
  type         = {Test Protocol},
  institution  = {European New Car Assessment Programme},
  note         = {Accessed: August 18, 2025}
}

@inproceedings{brannstrom2008situation,
  title={A situation and threat assessment algorithm for a rear-end collision avoidance system},
  author={Brannstrom, Mattias and Sjoberg, Jonas and Coelingh, Erik},
  booktitle={IV},
  pages={102--107},
  year={2008},
  organization={IEEE}
}

@article{lee1976theory,
  title={A theory of visual control of braking based on information about time-to-collision},
  author={Lee, David N},
  journal={Perception},
  volume={5},
  number={4},
  pages={437--459},
  year={1976},
  publisher={SAGE Publications Sage UK: London, England}
}

@article{razeena2025deep,
  title={Deep Learning based Automated Braking Decision-making for Advanced Driver Assistance System},
  author={Razeena, CP and Simon, Philomina},
  journal={Procedia Comput. Sci.},
  volume={258},
  pages={552--562},
  year={2025},
  publisher={Elsevier}
}

@inproceedings{isele2018safe,
  title={Safe reinforcement learning on autonomous vehicles},
  author={Isele, David and Nakhaei, Alireza and Fujimura, Kikuo},
  booktitle={IROS},
  pages={1--6},
  year={2018},
  organization={IEEE}
}

@inproceedings{chae2017autonomous,
  title={Autonomous braking system via deep reinforcement learning},
  author={Chae, Hyunmin and Kang, Chang Mook and Kim, ByeoungDo and Kim, Jaekyum and Chung, Chung Choo and Choi, Jun Won},
  booktitle={ITSC},
  pages={1--6},
  year={2017},
  organization={IEEE}
}

@inproceedings{han2014development,
  title={Development of autonomous emergency braking control system based on road friction},
  author={Han, I-Chun and Luan, Bi-Cheng and Hsieh, Feng-Chi},
  booktitle={CASE},
  pages={933--937},
  year={2014},
  organization={IEEE}
}

@article{zhang2024dual,
  title={Dual-AEB: Synergizing Rule-Based and Multimodal Large Language Models for Effective Emergency Braking},
  author={Zhang, Wei and Li, Pengfei and Wang, Junli and Sun, Bingchuan and Jin, Qihao and Bao, Guangjun and Rui, Shibo and Yu, Yang and Ding, Wenchao and Li, Peng and others},
  journal={arXiv preprint arXiv:2410.08616},
  year={2024}
}

@inproceedings{lee2013pseudo,
  title={Pseudo-label: The simple and efficient semi-supervised learning method for deep neural networks},
  author={Lee, Dong-Hyun and others},
  booktitle={ICML Workshop on Challenges in Representation Learning},
  volume={3},
  number={2},
  pages={896},
  year={2013},
  organization={Atlanta}
}

@inproceedings{xie2020self,
  title={Self-training with noisy student improves imagenet classification},
  author={Xie, Qizhe and Luong, Minh-Thang and Hovy, Eduard and Le, Quoc V},
  booktitle={CVPR},
  pages={10687--10698},
  year={2020}
}

@inproceedings{lee2024learning,
  title={Learning from Spatio-temporal Correlation for Semi-Supervised LiDAR Semantic Segmentation},
  author={Lee, Seungho and Lee, Hwijeong and Shim, Hyunjung},
  booktitle={IROS},
  pages={14095--14102},
  year={2024},
  organization={IEEE}
}

\end{document}